\documentclass[10pt,twocolumn,letterpaper]{article}

\usepackage{cvpr}
\usepackage{times}
\usepackage{epsfig}
\usepackage{graphicx}
\usepackage{amsmath}
\usepackage{amssymb}
\usepackage{paralist}
\usepackage[numbers,sort,compress]{natbib}

\usepackage[T1]{fontenc}
\usepackage[utf8]{inputenc}
\usepackage{authblk}

\usepackage[pagebackref=false,breaklinks=false,letterpaper=false,colorlinks,bookmarks=false]{hyperref}

\cvprfinalcopy 

\def\cvprPaperID{214} 

\graphicspath{{figures/images/}} 

\usepackage{comment}
\usepackage{color}
\definecolor{green}{rgb}{0, 0.5, 0}
\definecolor{orange}{rgb}{0.8, 0.6, 0.2}
\definecolor{red}{rgb}{1.0, 0.0, 0.0}
\definecolor{teal}{rgb}{0.0, 0.4, 0.4}
\definecolor{purple}{rgb}{0.65,0,0.65}
\definecolor{saffron}{rgb}{0.95,0.75,0.2}
\definecolor{turquoise}{rgb}{0.0,0.5,0.5}

\usepackage[ruled,vlined,linesnumbered]{algorithm2e}

\usepackage{multirow}

\usepackage[normalem]{ulem}

\usepackage{mathtools}
\usepackage{amsmath, amsthm, amssymb, amsfonts, amsopn}
\usepackage{graphicx} 
\usepackage{textcomp}
\usepackage{xfrac}
\usepackage{bbm}

\usepackage{overpic}
\usepackage{subfig}
\usepackage{wrapfig}

\newcommand{\hidecomment}[1]{}

\newcommand{\bz}{\mathbf{z}}

\usepackage{nicefrac}
\usepackage{cleveref}
\usepackage{booktabs}

\ifcvprfinal\fi
\begin{document}

\title{Hierarchy Denoising Recursive Autoencoders for 3D Scene Layout Prediction}

\author[1]{Yifei Shi}
\author[2]{Angel Xuan Chang}
\author[3]{Zhelun Wu}
\author[2]{Manolis Savva}
\author[1]{Kai Xu\thanks{corresponding author}}
\affil[1]{National University of Defense Technology}
\affil[2]{Simon Fraser University}
\affil[3]{Princeton University}

\renewcommand\Authands{ and }

\maketitle

\begin{abstract}

Indoor scenes exhibit rich hierarchical structure in 3D object layouts.
Many tasks in 3D scene understanding can benefit from reasoning jointly about the hierarchical context of a scene, and the identities of objects.
We present a variational denoising recursive autoencoder (VDRAE) that generates and iteratively refines a hierarchical representation of 3D object layouts, interleaving bottom-up encoding for context aggregation and top-down decoding for propagation.
We train our VDRAE on large-scale 3D scene datasets to predict both instance-level segmentations and a 3D object detections from an over-segmentation of an input point cloud.
We show that our VDRAE improves object detection performance on real-world 3D point cloud datasets compared to baselines from prior work.

%
\end{abstract}

\section{Introduction}
\label{sec:intro}


The role of \emph{context} in 3D scene understanding is central.
Much prior work has focused on leveraging contextual cues to improve performance on various perception tasks such as object categorization~\cite{galleguillos2010context}, semantic segmentation~\cite{mottaghi2014role}, and object relation graph inference from images~\cite{zellers2018scenegraphs}.
However, the benefit of hierarchical context priors in 3D object detection and 3D instance-level segmentation using deep learning is significantly less explored.
A key challenge in using deep network formulations for capturing the patterns of hierarchical object layout is that these patterns involve changing numbers of objects with varying semantic identities and relative positions.
In this paper, we propose a recursive autoencoder\footnote{Also known as a recursive neural network (RvNN) autoencoder.} (RAE) approach that is trained to predict and iteratively ``denoise'' a hierarchical 3D object layout for an entire scene, inducing 3D object detections and object instance segmentations on an input point cloud.

Recent work has demonstrated that encoding the context in 3D scene layouts using a set of pre-specified ``scene templates'' with restricted sets of present objects can lead to improvements in 3D scene layout estimation and 3D object detection~\cite{zhang2017deepcontext}.
However, manually specifying templates of scene layouts to capture the diversity of real 3D environments is a challenging and expensive endeavor.
Real environments are hierarchical: buildings contain rooms such as kitchens, rooms contain functional regions such as dining table arrangements, and functional regions contain arrangements of objects such as plates and cutlery.
This implies that explicit representation of the \emph{hierarchy structure} of 3D scenes can benefit 3D scene understanding tasks such as object detection and 3D layout prediction.


\begin{figure}
   \begin{overpic}[width=1.0\linewidth,tics=10]{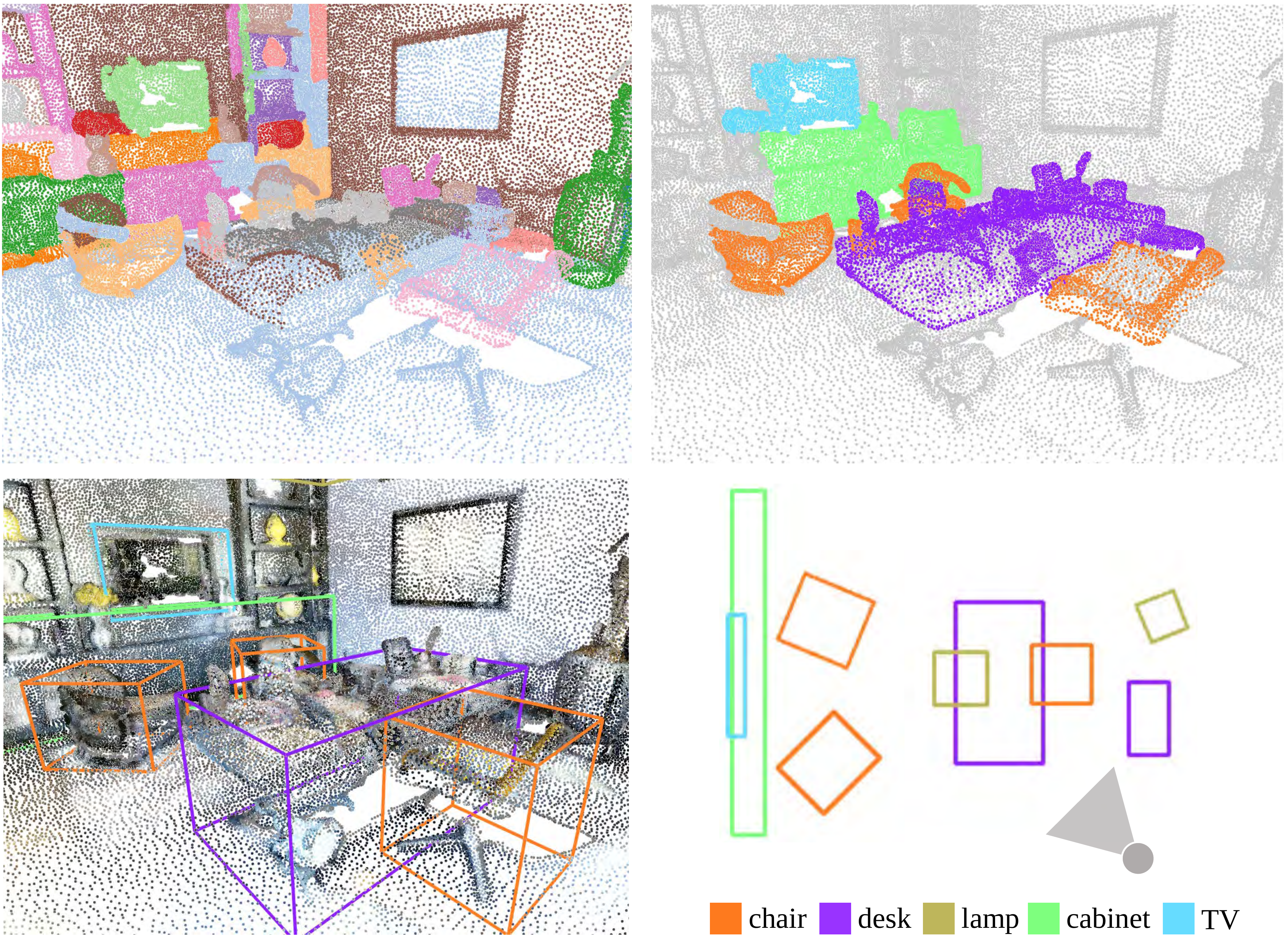}
   \end{overpic}
   \caption{
   We present a hierarchy-aware variational denoising recursive autoencoder (VDRAE) for predicting 3D object layouts.
   The input is a point cloud which we over-segment (top left).
   Our VDRAE constructs and refines a 3D object hierarchy, inducing semantic segmentations (top right, category-colored point cloud), and 3D instance oriented bounding boxes (bottom).
   The refined 3D bounding boxes tightly and fully contain observed objects.
   }
   \vspace{-1em}
   \label{fig:teaser}
   \vspace{-6pt}
\end{figure} 


Given a scene represented as a point cloud, we first perform an over-segmentation.
Our RvNN is trained to encode all segments in a \emph{bottom-up context aggregation} of per-segment information and inter-segment relations, forming a segment hierarchy.
The decoding phase performs a \emph{top-down context propagation} to regenerate subtrees of the hierarchy and generate object proposals.
This encoding-decoding refinement process is iterated, interleaving context aggregation and hierarchy refinement.
By training our denoising autoencoder in a generative fashion, this process converges to a refined set of object proposals whose layout lies in the manifold of valid scene layouts learned by the generative model (Figure~\ref{fig:teaser}).
In summary, our approach is an iterative \emph{3D object layout denoising autoencoder} which generates and refines object proposals by recursive context aggregation and propagation within the inferred hierarchy structure.
We make the following contributions:
\begin{itemize}
\vspace{-4pt}
  \item We predict and refine a multi-object 3D scene layout for an input point cloud, using hierarchical context aggregation and propagation based on a denoising recursive autoencoder (DRAE).
\vspace{-4pt}
  \item We learn a variational DRAE (VDRAE) to model the manifold of valid object layouts, thus facilitating layout optimization through an iterative infer-and-generate process.
\vspace{-4pt}
  \item We demonstrate that our approach improves 3D object detection performance on large reconstructed 3D indoor scene datasets. 
\end{itemize}


\begin{figure*}[t!] \centering
	\begin{overpic}[width=1.0\linewidth,tics=10]{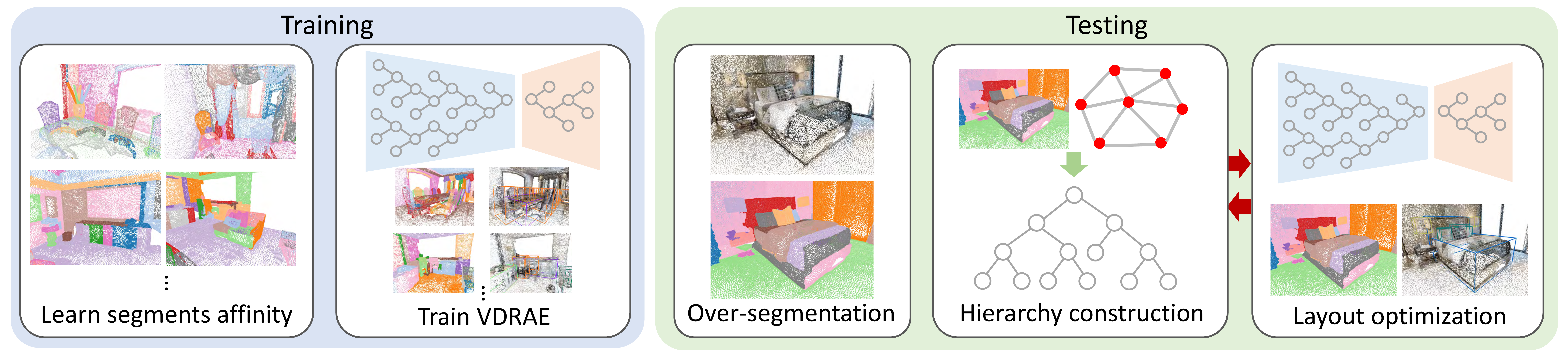}
   \end{overpic}
   \caption{
   Our system involves two neural net components: a segment-segment affinity prediction network which we use to construct hierarchical groupings of 3D objects, and a variational denoising recursive autoencoder (VDRAE) which refines the 3D object hierarchies.
   At test time, the affinity prediction network is used to predict segment-segment affinities.
   We construct a hierarchy from the segment affinity graph using normalized graph-cuts.
   The VDRAE then encodes this hierarchy to aggregate contextual queues and decodes it to propagate information between nodes.
   These two stages are iterated to produce a denoised set of 3D object detections and instance segmentations that better match the input scene.
   }
   \label{fig:overview}
\end{figure*} 

\section{Related work}
\label{sec:related}

Our goal is to improve 3D object detection by leveraging contextual information with a hierarchical representation of a 3D scene.
Here we focus on reviewing the most relevant work in object detection.
We describe prior work that uses context during object detection, work on object detection in 3D, and hierarchical context modeling.

\vspace{-1em}
\paragraph{Object detection in 2D.}
Object detection has long been recognized as an important problem in computer vision with much prior work in the 2D domain~\cite{felzenszwalb2010object,RCNNmR,RCNN,FastRCNN,redmon2016you,liu2016ssd,FasterRCNN,he2017mask}.
Using contextual information to improve object detection performance has also been studied extensively~\cite{choi2012tree,choi2012context,porway2008hierarchical,heitz2009cascaded,tu2005image}.
\citet{choi2013understanding} show that using contextual information enables predictions in 3D from RGB images.
More recently, \citet{zellers2018scenegraphs} show improved object detections by learning a global context using a scene graph representation.
These approaches operate in 2D and are subject to the viewpoint dependency of single image inputs.
The limited field of view and information loss during projection can significantly limit the benefit of contextual information.

\vspace{-1em}
\paragraph{Object detection with RGB-D.}
The availability of commodity RGB-D sensors led to significant advances in 3D bounding box detection from RGB-D image inputs~\cite{depthRCNN,SlidingShapes,DeepSlidingShapes,deng2017amodal}.
However, at test time, these object detection algorithms still only look at a localized region from a single view input and do not consider relationships between objects (i.e. contextual information).
There is another line of work that performs contextual reasoning on single view RGB-D image inputs~\cite{lin2013holistic,shao2014imagining,shi2016data,ren2016cog,zhang2017deepcontext,lahoud20172d,xu2018pointfusion} by leveraging patterns of multi-object primitives or point cloud segments to infer and classify small-scale 3D layouts.
\citet{zhang2017deepcontext} model rooms using four predefined templates (each defining a set of objects that may appear) to detect objects in RGB-D image inputs.
If the observed room contains objects that are not in the initial template, they cannot be detected.
Another line of work on street and urban RGB-D data uses bird's eye view representation to capture context for 3D object detection~\cite{simon2018complex,yang2018pixor,beltran2018birdnet}.
In contrast, we operate with fused 3D point cloud data of entire rooms, and learn a generative model of 3D scene layouts from a hierarchical representation.

\vspace{-1em}
\paragraph{Object detection in 3D point clouds.}
Recently, the availability of large scale datasets \cite{armeni20163d,dai2017scannet,chang2017matterport} has fostered advances in 3D scene understanding~\cite{xu2016data}.
There has been an explosion of  methods that focus on the semantic segmentation of point clouds~\cite{qi2016pointnet,tchapmi2017segcloud,engelmann2017spatial,li2018pointcnn,hua2018pointwise,landrieu2018superpoint,huang2018slice,ye20183d,lin2018multi}.
However, there is far less work addressing instance segmentation or object detection in fused 3D point clouds, at room-scale or larger.
Both \citet{qi2016pointnet,wang2018sgpn} propose connected component-based heuristic approaches to convert semantic segmentations to instances.
\citet{wang2018sgpn} is the state-of-the-art 3D point cloud instance segmentation method.
They use a learned point similarity as a proxy for context.
A related earlier line of work segments a point cloud or 3D mesh input into individual objects and then retrieves matching models from a CAD database~\cite{nan2012search,chen2014automatic,shao2014imagining,li2015database} to create a synthetic 3D representation of the input scene.
Our approach directly represents object detections as a hierarchy of 3D bounding boxes and is motivated by the observation that at the scale of 3D rooms, modeling the hierarchical context of the 3D object layout becomes important.

\vspace{-1em}
\paragraph{Hierarchical context in 3D.}
Hierarchical representations have been used to learn grammars in natural language and images~\cite{socher2011parsing}, 2D scenes~\cite{sharma2014recursive}, 3D shapes~\cite{yi17hierarchy,li2017grass}, and 3D scenes~\cite{liu2014creating}.
A related line of work parses RGB or RGB-D scenes hierarchically using And-Or graphs~\cite{zhao2011integrating,liu2018single,qi2018human,huang2018holistic,huang2018cooperative} for a variety of tasks.
For full 3D scenes, there has been very limited amount of available training data with ground truth hierarchy annotations.
Therefore, prior work in hierarchical parsing of 3D scenes does not utilize high capacity deep learning models.
For example, \citet{liu2014creating} is limited to training and testing on a few dozens of manually annotated scenes.
\citet{zhao2011image} evaluated on only 534 images.
In this paper, we use a recursive autoencoder neural network to learn a hierarchical representation for the entire 3D scene directly from large-scale scene datasets~\cite{chang2017matterport,armeni20163d}.

\section{Method}
\label{sec:method}

The input to our method is a 3D point cloud representing an indoor scene.
The output is a set $\mathcal{B}$ of objects represented by oriented bounding boxes (OBBs), each with a category label.
We start from an initial over-segmentation $\mathcal{S}$ containing candidate object parts (\Cref{sec:overseg}).
We then predict segment pair affinities and use a normalized cuts~\cite{shi2000normalized} approach to construct an initial hierarchy $h$ used for context propagation (\Cref{sec:hierarchy}).
Having built the hierarchy, we iteratively refine the 3D object detections and the hierarchy based on a recursive autoencoder network which adjusts the structure of the hierarchy and its nodes to produce 3D object detections at the leaf nodes (\Cref{sec:recursive}).
We call the combination of the object detections and the constructed hierarchy $\{\mathcal{B},h\}$ the 3D scene layout.
Our output set of labeled bounding boxes $\mathcal{B}$ contains a category label for all object detections, or a label indicating a particular box is not an object.
\Cref{fig:overview} shows an overview of our method.


\subsection{Initial Over-segmentation}
\label{sec:overseg}

Our input is a point cloud for which we create an initial over-segmentation $\mathcal{S}$ as a starting point for our object detection.
Distinct objects are represented by oriented bounding boxes containing parts of the point cloud.
We use features of the object points as well as features of the spatial relations between objects to characterize the object layout and to train our network such that it can detect objects.

There is much prior work that could be used to provide an initial over-segmentation of the point cloud.
We use a representative unsupervised method based on a greedy graph-based approach~\cite{felzenszwalb2010object} that was extended for point clouds by~\cite{karpathy2013object}.
Our method follows~\cite{karpathy2013object} in using graph cuts for an over-segmentation of the point cloud based on point normal differences to create the initial set of segments.

Each segment is extracted from the point cloud as an individual set of points, for which we compute oriented bounding boxes and point features as described in the following sections.


\subsection{Hierarchy Initialization}
\label{sec:hierarchy}

During hierarchy construction we address the following problem.
The input is the initial over-segmentation $\mathcal{S}$ and the output is a binary tree $h$ representing a hierarchical grouping of the objects.
Each object is represented as a 3D point cloud with an oriented bounding box (OBB), and a category label.
The 3D point cloud is a set of points $\{p_i\} = \{x_i, y_i, z_i, r_i, g_i, b_i\}$ with their 3D $(x,y,z)$ position and color $(r,g,b)$.
The leaves of this initial hierarchy $h$ are the segments and the internal nodes represent groupings of the segments into objects and groups of objects.
The root of the tree represents the entire room.


\begin{figure}[t!] \centering
	\begin{overpic}[width=1.0\linewidth,tics=10]{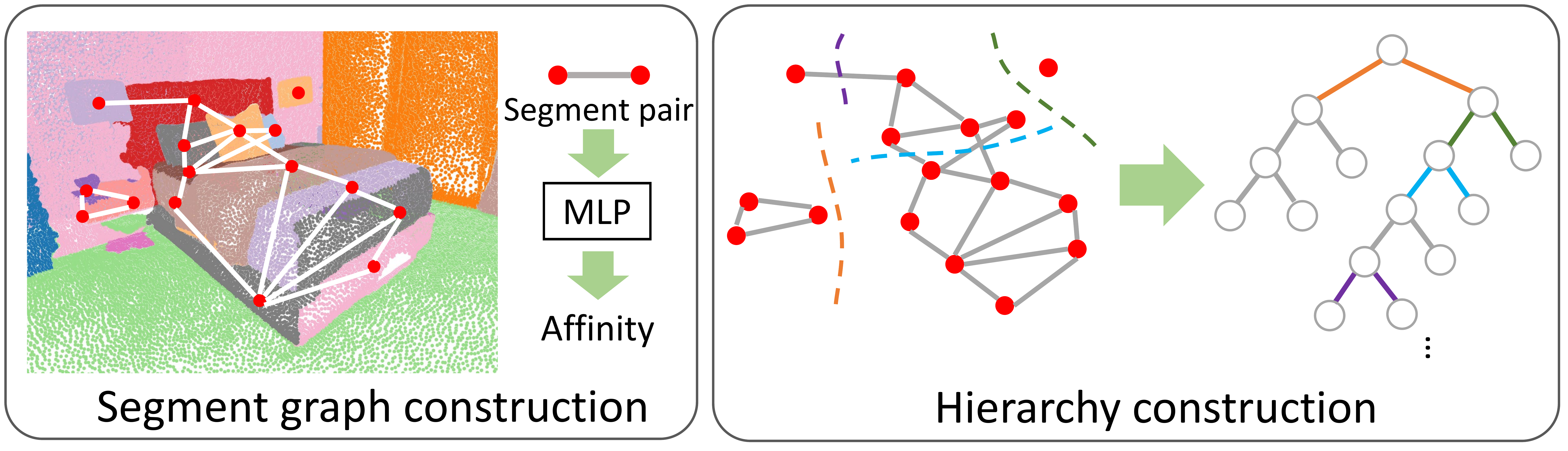}
    \put(20,-2){\small (a)}
    \put(68,-2){\small (b)}
	\end{overpic}
    \caption{
    We train an MLP to predict segment pair affinities and create a segment affinity graph \textbf{(a)}.
    We then construct a hierarchy from the resulting segment affinity graph using normalized cuts \textbf{(b)}.
    }
    \label{fig:hier}
    \vspace{-10pt}
\end{figure}

To construct the initial hierarchy from the input segments we first train a multi-layer perceptron (MLP) to predict segment pair affinities which indicate whether the two segments belong to the same object instance or not.
The input to the MLP is a set of features capturing segment-segment geometric and color relationships, proposed by prior work~\cite{xu2015autoscanning}.
We also tried using learned features obtained from a network trained on object-level label classification, but empirically found the features in~\cite{xu2015autoscanning} to work better in our experiments.
The MLP is trained to predict binary pair affinity from these features under a squared hinge loss.
Once we computed the segment pair affinities, the segments are then grouped into a hierarchy by using normalized cuts~\cite{shi2000normalized}.
Starting from the root node, we split the segments into two groups recursively.
The splitting stops when all groups have only one segment (leaf node).
The cut cost $E(u,v)=e_c e_a$ between two segments $u$ and $v$ in the normalized cut is initially equal to the affinity $e_a$ between the segments, but is then adjusted by the factor $e_c$ during layout optimization (as described in the next section). \Cref{fig:hier} shows the process of our hierarchy construction.


\subsection{Object Detection and Layout Refinement}
\label{sec:recursive}

We describe our iterative optimization for predicting the object layout $\{\mathcal{B},h\}$.
We begin with the basic recursive autoencoder (RAE) for context aggregation and propagation.
We then discuss a denoising version of the RAE (DRAE) designed for adjusting the object layout to better match a observed layout in the training set.
Based on that, we introduce a Variational DRAE (VDRAE) which is a generative model for object layout improvement.
It maps a layout onto a learned manifold of plausible layouts, and then generates an improved layout to better explain the input point cloud.

\begin{figure}[t!] \centering
	\begin{overpic}[width=1.0\linewidth,tics=10]{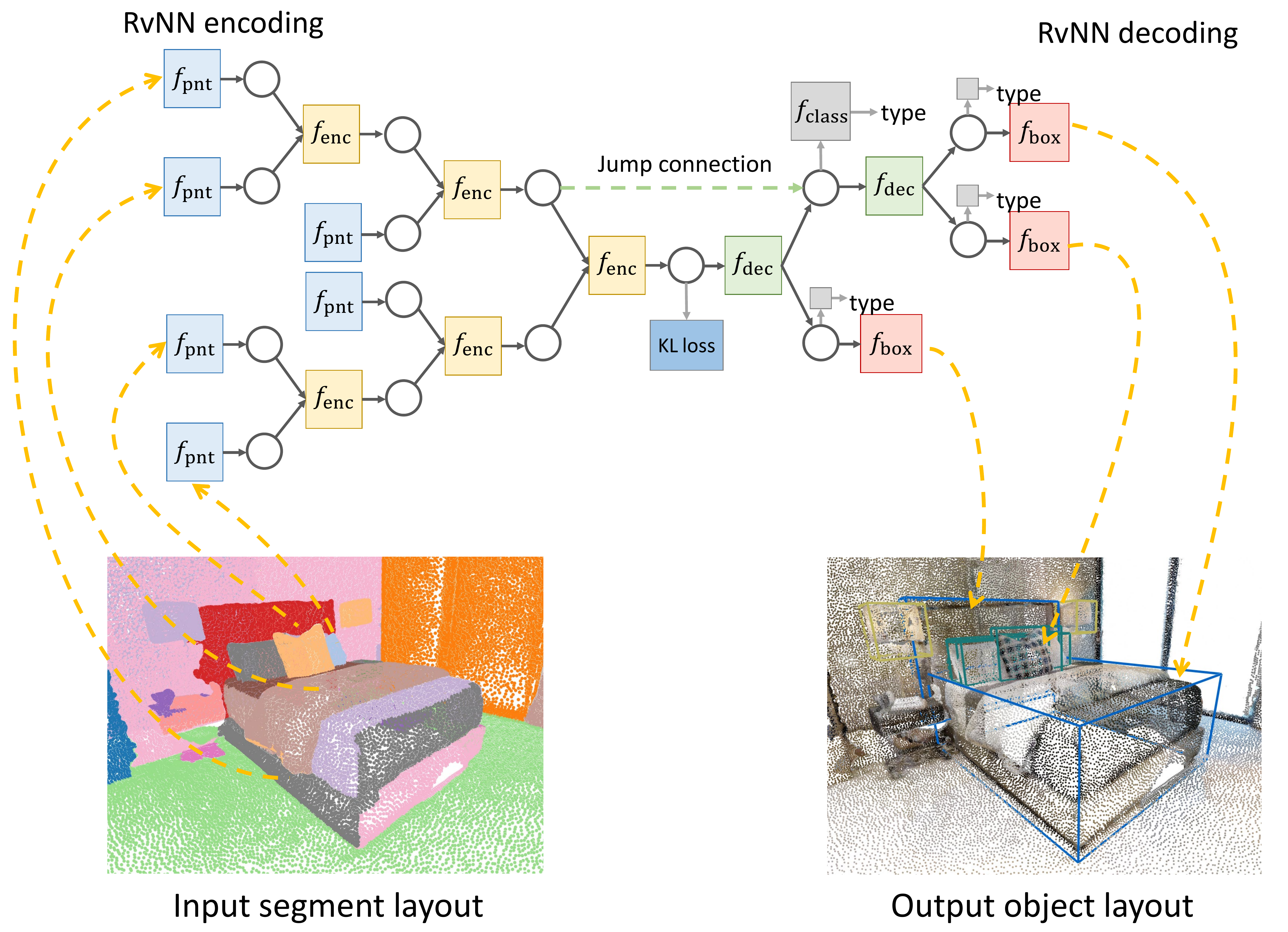}
	\end{overpic}
    \caption{
    Our denoising recursive autoencoder (RAE) takes an input segment layout from an over-segmentation and performs bottom-up encoding for context aggregation (left side), followed by top-down decoding for context propagation (right side).
    The encoding-decoding process generates a refined hierarchy with 3D object detections as leaf nodes.
    }
    \label{fig:rae}
    \vspace{-10pt}
\end{figure}

\paragraph{Recursive autoencoder for context propagation.}
Given the segments and the hierarchy, the recursive autoencoder (RAE) performs a bottom-up RvNN encoding for context aggregation, followed by a top-down RvNN decoding for context propagation.
The encoder network takes as input the features (codes) of any two nodes to be merged (according to the hierarchy) and outputs a merged code for their parent node: $x_p^{enc} = f_\text{enc}(x_l^{enc}, x_r^{enc})$, where $x_l^{enc}$, $x_r^{enc}$ and $x_p^{enc}$ denote the codes of two sibling nodes and their parent node, respectively.
$f_\text{enc}$ is a MLP with two hidden layers for node grouping.
The decoder takes the code of an internal node of the hierarchy as input and generates the codes of its two child nodes: $[x_l^{dec}, x_r^{dec}] = f_\text{dec}(x_p^{dec},x_p^{enc})$, where $f_\text{dec}$ is a two-layer MLP decoder for node ungrouping (Figure \ref{fig:rae}).


An additional {\em box encoder} generates the initial codes from the 3D point cloud within an OBB before the bottom-up pass, and a {\em box decoder} generates the final adjusted OBBs from the codes of the leaf nodes after the top-down pass: $x_n^{enc} = f_\text{pnt}(P_n), \quad t_n = f_\text{box}(x_n^{dec})$, where $x_n^{enc}$ and $x_n^{dec}$ denote the code for an node $n$ in encoding and decoding.
$P_n$ is the set of 3D points of node $n$.
$t_n$ is the parameter vector of an output OBB, encoding the offsets of its position, dimension and orientation.
$f_\text{pnt}$ is a PointCNN~\cite{li2018pointcnn} module for box encoding, and $f_\text{box}$ a two-layer MLP for box decoding.
The PointCNN\footnote{Alternative encoding modules such as PointNet++ can be used instead.} module is pretrained on a classification task for predicting object category labels from the point clouds of objects in the training set.

\paragraph{Denoising RAE for object detection and layout refinement.}
To endow the RAE with the ability to improve the predicted layout with respect to a target layout (e.g. a observed layout in the training set), a natural choice is to train a denoising RAE.
Given a noisy input segment layout, we learn a Denoising RAE (DRAE) which generates a denoised layout.
By noise, we mean perturbations over categorical labels, positions, dimensions and orientation of the bounding boxes.
In DRAE, denoising is accomplished by the decoding phase which generates a new hierarchy of OBBs that refines, adds or removes individual object OBBs.
The key for this generation lies in the node type classifier trained at each node (Figure \ref{fig:rae}) which determines whether a node is a leaf `object' node at which decoding terminates, or an internal node at which the decoding continues: $o_n = f_\text{cls}^\text{node}(x_n^{dec},x_n^{enc})$, with $o_n=0$ indicating a leaf `object' node and $o_n=1$ an internal `non-object' node
For 'object' nodes, another object classifier is applied to determine the semantic categories: $c_n = f_\text{cls}^\text{obj}(x_n^{dec}x_n^{enc})$, where $c_n$ is the categorical label for node $n$.
For training, we compute the IoU of all nodes in the encoding hierarchy against ground-truth object bounding boxes and mark all nodes with IoU $\leq 0.5$ as `object'.


\begin{algorithm}[t]\small
\caption{\mbox{VDRAE 3D Scene Layout Prediction.}}
\label{algo:rae}
\SetCommentSty{textsf}
\SetKwInOut{AlgoInput}{Input}
\SetKwInOut{AlgoOutput}{Output}
\SetKwFunction{ObjDetect}{ObjectDetection}
\SetKwFunction{OverSeg}{Over-segmentation}
\SetKwFunction{BuildHier}{HierarchyConstruction}
\SetKwFunction{TestVDRAE}{VDRAE}
\SetKwFunction{PartGeom}{PartGeometry}
\BlankLine
\AlgoInput{\mbox{Point cloud of indoor scene: $P$; Trained VDRAE.}}
\AlgoOutput{ 3D object layout $\{\mathcal{B},h\}$. }
$\mathcal{S}$ $\leftarrow$ \OverSeg{$P$}\;
$h$ $\leftarrow$ \BuildHier{$\mathcal{S}$, $P$}\;
\Repeat {Termination condition met} {
    $\mathcal{B}$ $\leftarrow$ \TestVDRAE{$\mathcal{S}$, $h$, $P$}\;
    $h$ $\leftarrow$ \BuildHier{$\mathcal{B}$, $\mathcal{S}$, $P$}\;
}
\Return $\{\mathcal{B},h\}$\;
\end{algorithm}

\paragraph{Variational DRAE for generative layout optimization.}
We train a Variational DRAE (VDRAE) to capture a manifold of valid hierarchies of OBBs from our training data.
The training loss is:
$$
\mathcal{L} = \sum\limits_{n}^\mathcal{N}{(\mathcal{L}^\text{node}_\text{cls}(o_n,o_n^*) + \mathcal{L}^\text{obj}_\text{cls}(c_n,c_n^*) +\mathcal{L}^\text{obj}_\text{obb}(t_n,t_n^*))} + \mathcal{L}_\text{KL}
$$
where $\mathcal{N}$ are all decoding nodes, $\mathcal{L}^\text{node}_\text{cls}$ is a binary cross-entropy loss over two categories (`object' vs `non-object'), $o_n^*$ is the ground-truth label, $\mathcal{L}^\text{obj}_\text{cls}$ is a multi-class cross-entropy loss over semantic categories, $o_n^*$ is the ground-truth categorical label, $\mathcal{L}^\text{obj}_\text{obb}$ is an $L_1$ regression loss on the OBB parameters of the node, $t_n^*$ is the ground-truth OBB parameters and $\mathcal{L}_\text{KL}$ is the KL divergence loss at root node.
Note that the $\mathcal{L}^\text{obj}_\text{cls}$ and $\mathcal{L}^\text{obj}_\text{obb}$ terms exist only for `object' nodes.
The last term serves as a regularizer which measures the KL divergence between the posterior and a normal distribution $p(\bz)$ at the root node.
This enables our VDRAE learning to map to the true posterior distribution of observed layouts.

\begin{figure}
\includegraphics[width=1.0\linewidth]{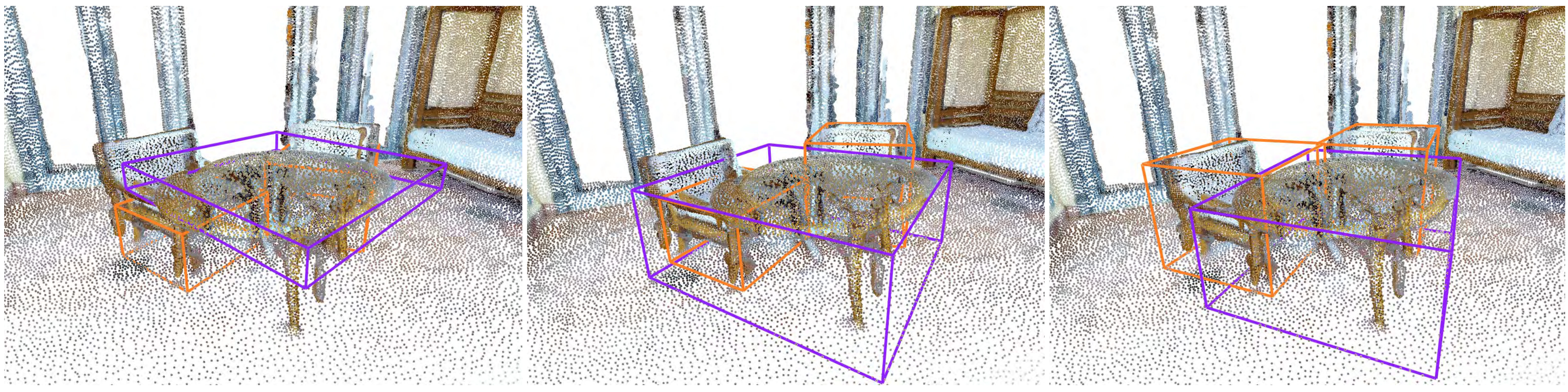}
\caption{
Example iterative refinement of initial object layout (leftmost column) under successive refinement iterations of our VDRAE network (columns to the right).
}
\label{fig:box_evolution}
\end{figure}

\paragraph{Layout refinement using the VDRAE.}
Once trained, the VDRAE can be used to improve an object layout.
Due to the coupling between object detection and hierarchy construction, we adopt an iterative optimization algorithm which alternates between the two stages (see \Cref{algo:rae}).
Given an initial segment layout extracted from the input point cloud, our method first performs a VDRAE inference step (test-time step) to generate a hierarchy of object bounding boxes explaining the input point cloud.
It then uses the decoding feature to infer a new hierarchy, which will be used for the VDRAE test in the next iteration. In the next iteration, the binary classification `object' vs `non-object' confidence is used to scale the normalized cut affinity $e_a$ for two nodes $u$ and $v$ using the following factor $e_c$:
$$
e_c(u,v) = \left\{\begin{array}{rr}
  -\text{log}(1-c_s), & u \text{ and } v \text{ in same leaf node $s$} \\
  0.1, & \text{otherwise}
\end{array}\right.
$$
where $c_s$ is the classification confidence of node $s$ to be labeled as `object'.
The scaled affinity $E(u,v)=e_c e_a$ is then used to refine the hierarchy construction.
This process repeats until the structure of the hierarchy between iterations remains unchanged.
\Cref{fig:box_evolution} shows an example of the iterative refinement.
The optimization converges with at most $5$ iterations for all the scenes we have tested.
This iterative optimization gradually ``pushes'' the object layout into the layout manifold learned by VDRAE. Please refer to the supplemental material for a discussion of convergence.


\section{Implementation Details}
\label{sec:impl}

In this section, we describe the implementation details of our network architectures, the relevant parameters, and the training and testing procedures.

\paragraph{Initial over-segmentation and feature extraction.}
For the initial over-segmentation we use threshold values $k=0.01$ which we empirically found to perform well on training scenes (\Cref{sec:overseg}).
For the PointCNN \cite{li2018pointcnn} features, we train the PointCNN to predict object class labels using the training set data.
We train the network to minimize the cross-entropy loss over $41$ object classes taking $2048$ points per input and outputting to a $256$-d vector for classification. Note that PointCNN is a pre-trained feature extractor and we didn't fine-tune it during the training of VDRAE.

\paragraph{Hierarchy construction.}
The MLP for segment pair affinity prediction consists of $4$ FC layers (with sigmoid layers).
The input is a $25$-d feature, and the output is a single affinity value.
We use the detault parameter setting for the solver used in normalized cuts.
It takes about $0.1s$ to build a hierarchy from a segment graph.


\paragraph{Variational denoising recursive autoencoder.}
$f_{enc}$ has two $1000$-d inputs and one $1000$-d output. $f_{dec}$ has one $1000$-d input and two $1000$-d outputs. $f_\text{cls}^\text{node}$ takes a $1000$-d vector as input, and outputs a binary label. $f_\text{cls}^\text{obj}$ takes a $1000$-d vector as input, and outputs a categorical label and OBB parameter offsets. This is achieved by using a softmax layer and a fully-connected layer.
To deal with the large imbalance between positive (`object') and negative (`non-object') classes during training, we use a focal loss~\cite{lin2018focal} with $\gamma=0$ for positives and $\gamma=5$ for negative. This makes the training focus on all positive samples and \emph{hard negative} samples.
All the items in $\mathcal{L}$ can be trained jointly. However, to make the training easier, we first train by $\mathcal{L}^\text{node}_\text{cls}$ and $\mathcal{L}_\text{KL}$ to make the network have the ability to distinguish whether a node is a single object, and then fine-tune by $\mathcal{L}^\text{obj}_\text{cls}$ and $\mathcal{L}^\text{obj}_\text{obb}$.

\paragraph{Training and testing details.}
We implement the segment pair affinity network and the VDRAE using PyTorch~\cite{paszke2017automatic}.
For VDRAE, We use the Adam optimizer with a base learning rate of $0.001$.
We use the default hyper-parameters of $\beta_1=0.9,\beta_2=0.999$ and no weight decay.
The batch size is $8$. The VDRAE can be trained in $15$ hours on a Nvidia Tesla K40 GPU.
At testing time, a forward pass of the VDRAE takes about 1s.
An Non-Maximum Suppression with IOU 0.5 is performed on the detected boxes after the inference of VDRAE. 


\begin{figure*}
\includegraphics[width=\linewidth]{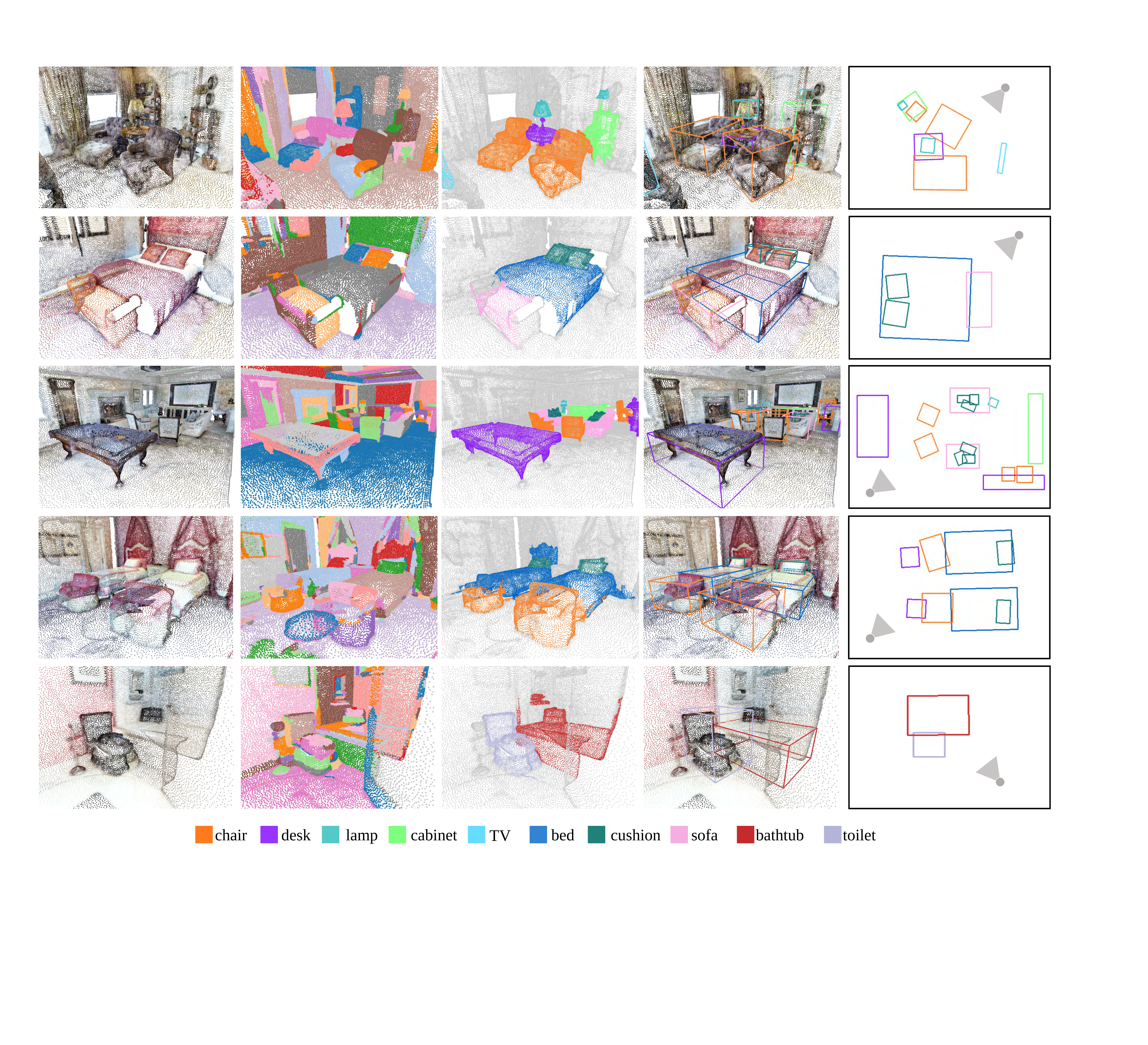}
\caption{
3D scene layout predictions using our VDRAE on the Matterport3D test set.
The first column shows the input point cloud.
The second column is the over-segmentation from which we construct an initial segment hierarchy.
The third column shows the 3D object detections with colors by category.
The final two columns show bounding boxes for the detections.
Our approach predicts hierarchically consistent 3D layouts where objects such as lamps, pillows and cabinets are detected in plausible positions and orientations relative to other objects and the global structure of the scene.
}
\label{fig:qualitative}
\end{figure*}

\section{Results}
\label{sec:result}

We evaluate our proposed VDRAE on 3D object detections in 3D point cloud scenes (see supplemental for semantic segmentation evaluation).

\subsection{Experimental Datasets}

We use two RGB-D datasets that provide 3D point clouds of interior scenes: S3DIS~\cite{armeni20163d} and Matterport3D~\cite{chang2017matterport}.
\emph{S3DIS} consists of six large-scale indoor areas reconstructed with the Matterport Pro Camera from three different university buildings.
These areas were annotated into $270$ disjoint spaces (rooms or distinct regions).
We use the k-fold cross validation strategy in~\cite{armeni20163d} for train and test.
\emph{Matterport3D} consists of semantically annotated 3D reconstructions based on RGB-D images captured from $90$ properties with a Matterport Pro Camera.
The properties are divided into room-like regions.
We follow the train/test split established by the original dataset, with $1,561$ rooms in the training set and $408$ rooms in the testing set.


\subsection{Evaluation}

Our main evaluation metric is the average precision of the detected object bounding boxes against the ground truth bounding boxes at a threshold IoU of $0.5$ (i.e. any detected bounding box that has more than $0.5$ intersection-over-union overlap with its ground truth bounding box is considered a match).
We compare our method against baselines from prior work on object detection in 3D point clouds.
We then present ablated versions of our method to demonstrate the impact of different components on detection performance, as well as experiments to analyze the impact of the over-segmentation coarseness and the impact of successive refinement iterations.

\paragraph{Qualitative examples.}
\Cref{fig:qualitative} shows detection results on the Matterport3D test set (see supplement for more examples).
Our VDRAE leverages hierarchical context to detect and refine 3D bounding boxes for challenging cases such as pillows on beds, and lamps on nightstand cabinets.


\begin{figure}
\begin{overpic}[width=1.0\linewidth,tics=10]{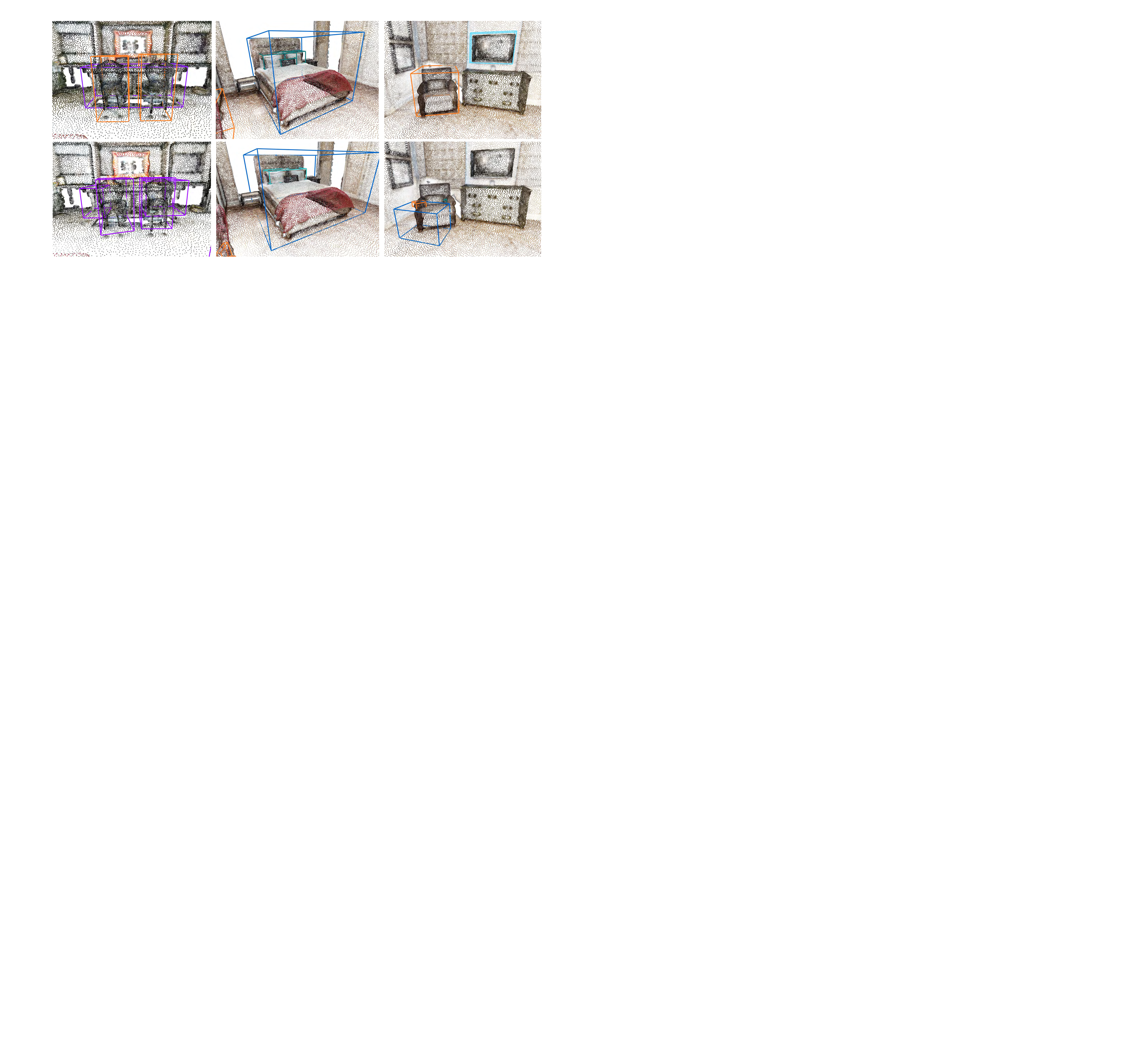}
\end{overpic}
\caption{
Qualitative 3D object detection results on Matterport3D test set using our VDRAE (top row) and the best performing baseline from prior work (SGPN\cite{wang2018sgpn}, bottom row).
Our approach produces more accurate bounding box detections and fewer category errors.
For example, chairs are correctly categorized and have tight bounding boxes at the top left and top right.
}
\label{fig:qualitative_compare}
\end{figure}

\paragraph{Comparison to baseline methods.}
We evaluate our approach against several baselines from prior work that produce object detections for indoor 3D scene point clouds:
\begin{compactitem}
\item \textbf{Seg-Cluster}: Approach proposed by~\cite{wang2018sgpn} applies semantic segmentation (SegCloud \cite{tchapmi2017segcloud}) followed by Euclidean clustering~\cite{rusu2011pcl}.
\item \textbf{PointNet}: Predicts the category of points~\cite{qi2016pointnet} and uses breadth-first search to group nearby points with the same category, inducing object instances. We use PointNet instead of other point-based neural networks as PointNet proposed this object detection pipeline.
\item \textbf{Sliding PointCNN}: A baseline using a 3D sliding window approach with PointCNN~\cite{li2018pointcnn} features.
\item \textbf{SGPN}: A state-of-the-art semantic instance segmentation approach for point clouds~\cite{wang2018sgpn}.
\item \textbf{Ours (flat context)}: A baseline using a flat context representation instead of leveraging the hierarchy structure, in which $x_n^{dec}$ is the concatenation of encoded features $x_n^{enc}$ and the average encoded features of all nodes $(\sum\nolimits_{n}^\mathcal{N}{x_n^{enc}})/n$.
\end{compactitem}
\Cref{tab:s3dis_results,tab:matterport_results} report average precision on the S3DIS and Matterport3D datasets, showing that our approach outperforms all baselines.
The flat context baseline performs worse than our hierarchy-aware VDRAE but better than the baselines that do not explicitly represent context.
\Cref{fig:qualitative_compare} qualitatively shows results from the Matterport3D test set, comparing our approach with the highest performing prior work baseline using SGPN.

\begin{table}
\footnotesize\centering
\begin{tabular}{lccccc}
\toprule
 & Chair & Table & Sofa & Board & mAP \\
\midrule
Seg-Cluster~\cite{wang2018sgpn} & 0.23 & 0.33 & 0.05 & 0.13 & 0.19 \\
Sliding PointCNN~\cite{li2018pointcnn}              & 0.36 & 0.39 & 0.23 & 0.07 & 0.26 \\
PointNet~\cite{qi2016pointnet}         & 0.34 & 0.47 & 0.05 & 0.12 & 0.25 \\
SGPN~\cite{wang2018sgpn}               & 0.41 & 0.50 & 0.07 & 0.13 & 0.28 \\
\midrule
Ours (flat context) & 0.35 & 0.47 & 0.32 & 0.10 & 0.31 \\
Ours & \textbf{0.45} & \textbf{0.53} & \textbf{0.43} & \textbf{0.14} & \textbf{0.39} \\
\bottomrule
\end{tabular}
\caption{
Comparison of our approach against prior work on object detection in 3D point cloud data.
Values report average precision at IOU of $0.5$ on the S3DIS dataset.
Our hierarchy-refining VDRAE outperforms all prior methods.
}
\label{tab:s3dis_results}
\end{table}

\begin{table*}
\footnotesize\centering
\begin{tabular*}{\linewidth}{@{\hskip 6pt\extracolsep{\stretch{1}}}*{13}{c}*{1}{c}}
\toprule
 & Chair & Table & Cabinet & Cushion & Sofa & Bed & Sink & Toilet & TV & Bathtub & Lighting & mAP \\
\midrule
Sliding PointCNN~\cite{li2018pointcnn} & 0.22 & 0.21 & 0.03 & 0.19 & 0.20 & 0.36 & 0.07 & 0.16 & 0.05 & 0.15 & 0.10 & 0.16 \\
PointNet~\cite{qi2016pointnet} & 0.28 & \textbf{0.32} & 0.06 & 0.21 & 0.28 & 0.25 & 0.17 & 0.08 & 0.10 & 0.11 & 0.06 & 0.18\\
SGPN~\cite{wang2018sgpn} & 0.29 & 0.24 & 0.07 & 0.18 & \textbf{0.30} & 0.33 & 0.15 & 0.17 & 0.09 & 0.16 & 0.11 & 0.19 \\
\midrule
Ours (flat context) & 0.24 & 0.18 & 0.08 & 0.21 & 0.18 & 0.27 & 0.22 & 0.25 & 0.07 & 0.21 & 0.07 & 0.18 \\
Ours & \textbf{0.37} & 0.27 & \textbf{0.11} & \textbf{0.24} & 0.28 & \textbf{0.43} & \textbf{0.23} & \textbf{0.35} & \textbf{0.19} & \textbf{0.27} & \textbf{0.19} & \textbf{0.27} \\
\bottomrule
\end{tabular*}
\caption{
Average precision of object detection at IoU $0.5$ on the Matterport3D dataset.
We compare our full method (`ours') against several baselines.
Refer to text for the details of the baselines.
}
\label{tab:matterport_results}
\end{table*}

\paragraph{Ablation of method components.}
We evaluate the impact of each components using the following variants:
\begin{compactitem}
\item \textbf{No hierarchy}: We use PointCNN~\cite{li2018pointcnn} features for each node to predict the object category and regress an OBB without using a hierarchy. We add $4$ FC layers after the PointCNN layers to increase the number of network parameter and make the comparison fair.
\item \textbf{No OBB regression}: We turn off the OBB regression module for leaf nodes and train from scratch.
\item \textbf{No iteration (bvh)}: No iteration for testing. The hierarchy is constructed through recursive binary splits considering only geometric separation between segments, i.e. bounding volume hierarchy (bvh).
\item \textbf{No iteration (our hier)}: No iteration for testing. The hierarchy is built by our hierarchy initialization approach.
\end{compactitem}

\Cref{tab:s3dis_ablation} shows the results.
The full method performs the best.
Not using a hierarchy degrades performance the most.
Removing OBB regression, and not performing iterative refinement are also detrimental but to a lesser extent.

\begin{table}
\footnotesize\centering
\begin{tabular}{lccccc}
\toprule
 & Chair & Table & Sofa & Board & mAP \\
\midrule
no hierarchy & 0.34 & 0.41 & 0.35 & 0.08 & 0.30 \\
no OBB regression & 0.41 & 0.47 & 0.40 & 0.11 & 0.35 \\
no iteration (bvh) & 0.37 & 0.47 & 0.38 & 0.10 & 0.33 \\
no iteration (our hier) & 0.39 & 0.51 & 0.39 & 0.12 & 0.35 \\
\midrule
Ours & \textbf{0.45} & \textbf{0.53} & \textbf{0.43} & \textbf{0.14} & \textbf{0.39} \\
\bottomrule
\end{tabular}
\caption{
Ablation of the components of our approach.
Values report average precision at IoU of $0.5$ on the S3DIS dataset.
Our full VDRAE outperforms all ablations.
}
\label{tab:s3dis_ablation}
\end{table}

\begin{figure}
\begin{overpic}[width=1.0\linewidth,tics=10]{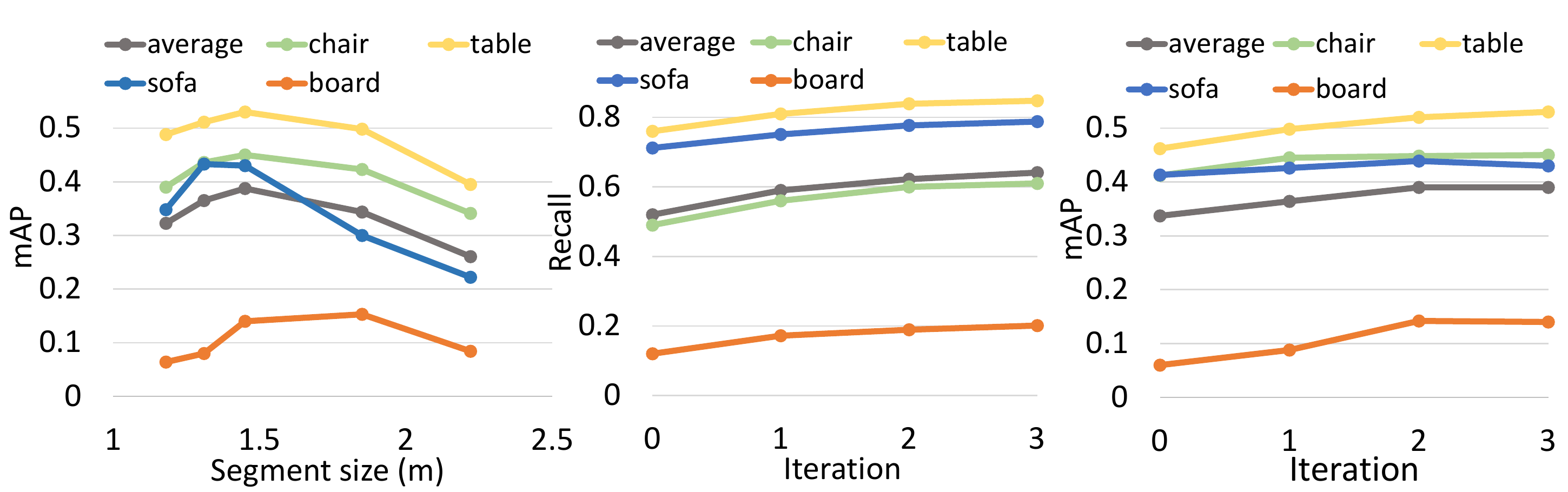}
\put(15,31){\small (a)}
\put(50,31){\small (b)}
\put(85,31){\small (c)}
\end{overpic}
\caption{
(a) mAP plotted against over-segmentation coarseness (average segment size in meters).
(b) recall against VDRAE iteration count.
(c) mAP against VDRAE iteration count.
}
\label{fig:iteration_coarseness}
\vspace{-10pt}
\end{figure}

\paragraph{Sensitivity to over-segmentation coarseness.}
We quantify the impact of the over-segmentation coarseness threshold parameter $k$ of the method in~\cite{karpathy2013object} on S3DIS.
We use five threshold values $k=1.0,0.1,0.01,0.001,0.0001$ to generate segments with different size and re-train the affinity network and VDRAE respectively.
Larger $k$ produce bigger segments.
\Cref{fig:iteration_coarseness} (a) shows that the best performance is achieved when average segment size is $1.45\text{m}$ ($k=0.01$).

\paragraph{Effect of iteration.}
We evaluate the effect of VDRAE refinement iteration by analyzing the hierarchy and 3D object detections at each step. \Cref{fig:iteration_coarseness} (b) shows recall against ground-truth objects plotted against iteration number.
Recall is computed by calculating the IoU of the OBB of each ground-truth object with \emph{all node OBBs in encoding hierarchy}. If one of the IoU values is larger than $0.5$, we consider that a match against the ground-truth. 
\Cref{fig:iteration_coarseness} (c) shows the object detection mAP plotted against iteration number.
The benefit of iteration is apparent.


\section{Conclusion}
\label{sec:future}

We presented an approach for predicting 3D scene layout in fused point clouds by leveraging a hierarchical encoding of the context.
We train a network to predict segment-to-segment affinities and use it to propose an initial segment hierarchy.
We then use a variational denoising recursive autoencoder to iteratively refine the hierarchy and produce 3D object detections.
We show significant improvements in 3D object detection relative to baselines taken from prior work.

\paragraph{Limitations.}
Our current method has several limitations.
First, the hierarchy proposal and VDRAE are trained separately.
Incorporating these two stages will leverage the synergy between parsing hierarchies and refining the 3D scene layouts.
Second, the segment point features we use in our VDRAE is trained independently on a classification task.
These features can also be learned end-to-end, resulting in further task-specific improvements in performance.

\paragraph{Future work.}
We have only taken a small step towards leveraging hierarchical representations of 3D scenes.
There are many avenues to pursue for future research.
Reasoning about the hierarchical composition of scenes into objects, object groups, functional regions, rooms, and entire residences can benefit many tasks beyond 3D object detection.
We hope that our work will act as a catalyst in this promising research direction.

\section*{Acknowledgements}
We are grateful to Thomas Funkhouser and Shuran Song for the valuable discussion.
This work was supported in part by NSFC (61572507, 61532003, 61622212) and Natural Science Foundation of Hunan Province for Distinguished Young Scientists (2017JJ1002).

{\small
\bibliographystyle{plainnat}
\setlength{\bibsep}{0pt}
\bibliographystyle{abbrvnat}
\bibliography{sceneparse}
}

\end{document}